# Multi-level Binarized LSTM in EEG Classification for Wearable Devices

Najmeh Nazari*, Seyed Ahmad Mirsalari*, Sima Sinaei+, Mostafa E. Salehi*, Masoud Daneshtalab+
*School of Electrical and Computer Engineering, University of Tehran, Tehran, Iran
+Division of Intelligent Future Technologies, Malardalen University, Vasteras, Sweden
{najme.nazari, ahmad.mirsalari, mersali}@ut.ac.ir, {sima.sinaei, masoud.daneshtalab}@mdh.se

*Abstract*— Long Short-Term Memory (LSTM) is widely used in various sequential applications. Complex LSTMs could be hardly deployed on wearable and resourced-limited devices due to the huge amount of computations and memory requirements. Binary LSTMs are introduced to cope with this problem, however, they lead to significant accuracy loss in some applications such as EEG classification which is essential to be deployed in wearable devices. In this paper, we propose an efficient multi-level binarized LSTM which has significantly reduced computations whereas ensuring an accuracy pretty close to full precision LSTM. By deploying 5-level binarized weights and inputs, our method reduces area and delay of MAC operation about 31× and 27× in 65nm technology, respectively with less than 0.01% accuracy loss. In contrast to many compute-intensive deep-learning approaches, the proposed algorithm is lightweight, and therefore, brings performance efficiency with accurate LSTM-based EEG classification to real-time wearable devices.

## I. INTRODUCTION

Deep Neural Networks (DNNs) have recently attained state of the art results in many machine learning applications such as natural language processing [2], computer vision [3], speech recognition [4], and medical diagnoses [5]. One of the important subsets of machine learning tasks is sequence learning that deals with data for which order is important. In these tasks, the data is arranged in a specific order, and the specified order is relevant to the task at hand. There is a wide range of sequence learning problems, for which particular classes of neural networks, called Recurrent Neural Networks (RNN), often yield state-of-the-art results. One of the most popular and effective RNN models used for sequence learning is called Long Short-Term Memory (LSTM) [1]. LSTM was designed to model the long-range dependencies and memory backup of RNN plays a very important role and so they are turned to be more accurate and effective than conventional RNNs. LSTM has been shown to provide state-of-the-art performance in various sequence learning tasks, including health care application and medical diagnoses such as ElectroEncephaloGraphy (EEG) [5] and Electrocardiogram (ECG) [6] signal classification.

LSTMs have commonly stand massive computations and the enormous amount of memory accesses, which are the significant challenges of them. As LSTMs go more extensive, they ensure higher accuracy, but at the cost of higher complexity. An LSTM model must be computed on hundreds of megabytes worth of parameters. Therefore, in addition to on-chip static random-access memory (SRAM), deploying off-chip dynamic random-access memory (DRAM) is essential for storing these data in hardware architecture. It should be noted that two orders of magnitude of the energy budget of an embedded device are consumed by off-chip memory accesses compared to on-chip memory accesses which increases the overall system latency significantly [7]. In other words, the off-chip parameter access can also be a processing bottleneck because the speed of matrix-vector operations is often limited by the memory access time [18]. Indeed, in LSTMs, like many other neural network models, the majority of computations are matrix-vector operations and have their parameter in the off-chip memory, so memory bandwidth profoundly affects system throughput.

Due to the growing demands of running high-accurate and yet energy-efficient LSTMs on battery-operated and resource-limited embedded devices, deploying efficient models in a hardware-friendly manner is essential. There are some solutions for efficient hardware implementation of LSTM network inference as follow: I. Model compression by reducing the amount of data needed for computations: it also determines the required computational elements, the memory requirements, and ultimately the performance and throughput of the system [8]. II. Memory storage optimization: Optimizing the memory accesses is positively correlated with embedded system performance [9][10].

Compacting neural network architectures and pruning techniques are used to decrease the volume of operations and the model size. Many prior works have substituted floating-point with fixed-point operands to decrease computation complexity [11][12]. Although they have significantly reduced computations, they still consist of computation-intensive MAC operations. Binary neural networks (BNNs) have tried to reduce MAC operation by constraining operand to {-1, 1} and hence MAC operations are substitute with XNOR bitcount operation [13][14]. Although BNNs decrease the huge amount of computations, they suffer from accuracy loss. Ternary neural networks have attempted to improve accuracy by adding an extra value to binary value ({-1,0,1}) whereas reduce MAC operation similar to BNNs. TOT-Net as a remarkable TNNs has deployed XOR and AND gate instead of multiplication and also achieved more accuracy compared to BNNs [23]. BNNs and TNNs are extremely suitable for real-time and battery-operated embedded systems and wearable devices. However, deploying the BNN model to large datasets, such as EEG, leads to remarkable accuracy loss. In this paper, many efforts are deployed to cope with accuracy loss while still reduce computation complexity.

This paper focuses on the LSTM model compression and is concerned with the design of a high-performance and energy-efficient solution for implementing deep learning inference. We have particularly focused on datasets that have remarkable accuracy loss by binarized LSTM. Therefore, we have introduced a simple and efficient representation for both weights and inputs of LSTM to decrease the accuracy loss and achieve a high accuracy which is pretty close to a full precision LSTM. Our proposed method significantly reduces computations, area, and energy consumption yet with an extremely negligible accuracy loss compared to the full-precision LSTM. Based on application demand, our proposed method has provided a trade-off between accuracy and volume of computation which is arisen from the input and weight bit-width. To put it simply, if accuracy is crucial for application, we can use more bit-widths for weights and inputs to achieve higher accuracy tremendously close to full precision accuracy. The rest of the paper is organized as follows. In Section II, we briefly overview related work. Section III presents a background on the LSTM. Our proposed method is discussed in Section IV. Section V presents the experimental results and finally, the paper is concluded in Section VI.

## II. RELATED WORK

Several previous studies have addressed the problem of implementing LSTM network inference in a hardware-friendly manner. As mentioned before, complete solutions to this problem should address three issues: first, reduce the amount of data and parameters by compressing the model, then optimize the dataflow for the target hardware platform and finally, optimize the memory usage.

Reducing the amount of data and parameters is one of the most important steps in optimizing the hardware implementation of LSTM. It would be obtained by compressing the model and selecting the data format and bit-precision of the numbers and a model compression strategy. This step determines the required computational elements, the memory requirements, and ultimately the performance and throughput of the system. Parameter compression can significantly reduce the memory footprint and computational complexity of the inference computations. Several methods have been proposed for neural network model compression. These methods are generally classified into parameter reduction and parameter quantization approaches. The former approach reduces and prunes the number of parameters needed to run the model, while the latter reduces the number of bits needed to represent the model parameters. In this paper, we focus on parameter compression. This section provides an overview of academic literature in model and parameter compression of deep learning algorithms for embedded systems.

The *pruning* technique reduces the number of parameters by removing (i.e. setting to zero) unnecessary weight connections. Using this technique, Han et al. demonstrated 9× and 13× reduction in the number of parameters for two well-known convolutional networks [15]. Building upon this approach, Han et al. also proposed a hardware-friendly pruning method, called load-balance aware pruning [16]. This method facilitates the distribution of the processing workload by constraining regions of the parameter matrices to contain an equal number of non-zero connections.

*Singular Value Decomposition* (SVD) is another technique for reducing the number of model parameters. By keeping the largest singular values, the dimensionality of model parameter matrices can be reduced while maintaining an acceptable level of accuracy. In [18], SVD is applied to a fully-connected layer of a convolutional network and achieved a compression rate of 7× with only 0.04% loss in prediction accuracy. In [19], this technique is applied to a 5-layer LSTM for speech processing and reduced the model size by 3× with only 0.5% loss in prediction accuracy.

A common technique for quantizing the model parameters is *weight sharing*. With this approach, small number of effective weight values are used [15]. After training, similar weights are identified using k-means clustering. The centroid of clusters is then chosen to be the shared weight values. The quantized model can then be retrained using only the shared weights. This technique allows parameter matrices to be stored as indices into the shared weight codebook, thus reducing the memory footprint.

Another quantization technique is called *binarization*. This method constrains all weight values to +1 or -1 during training and exploits a single bit for representing these two values. An extension of this technique applies the same quantization on activation values as well. Various approaches have been proposed for training a convolutional neural network with binary weights and activations [21][22]. The approach proposed in [22] was applied to an LSTM model for text prediction and resulted in better power and performance efficiency with negligible loss in prediction accuracy, compared to its full-precision model. Binarization results in significant advantages in efficient hardware implementation. In a weight-only binarization scheme, multiplications are reduced to multiplexer operation, thus eliminates hardware-expensive multipliers. Binarization of both weights and activations further simplifies hardware implementation, since computation-intensive multiply-accumulations are replaced with simple XNOR and bit-count operations. In both schemes, the size of the model parameter storage is drastically reduced. In this paper, we substitute full-precision weights and activations with multi-level binarized weights and activations to significantly reduce computations yet retain accuracy close to full-precision LSTM.

## III. BACKGROUND

### A. RNN cell Types

*Simple RNN Cell:* Figure 1(a) shows a simple RNN cell. Where $x_t$ is the input vector at time $t$. $h_t$ and $c_t$ are state vectors which are carried from time $t-1$ to time $t$, and hence, act as a memory by encoding previous information. $h_t$ is also considered as the cell output. Size of $h$ and $c$ vectors are denoted by $N_h$ and is known as the number of hidden units. The cell works based on the following equations.

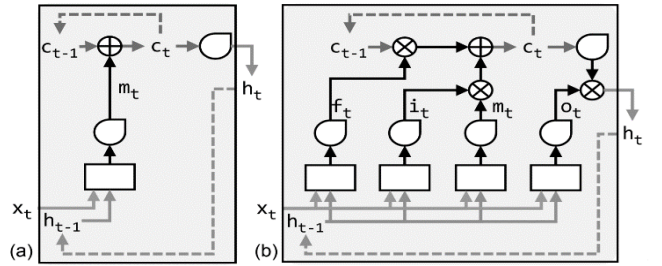

Figure 1. (a) Simple RNN Cell, (b) Long Short-Term Memory (LSTM)

$$m_t[j] = \tanh\left(\begin{array}{c}\sum_{k=1}^{N_x} w[j,k] x_t[k] + \\ \sum_{k=1}^{N_h} u[j,k] h_{t-1}[k] + b[j]\end{array}\right) \quad (1)$$

$$c_t[j] = c_{t-1}[j] + m_t[j] \quad (2)$$

$$h_t[j] = \tanh(c_t[j]) \quad (3)$$

As shown in (1), an intermediate vector $m_t$ is formed by applying *tanh* activation function on a linear combination of $x_t$ and $h_{t−1}$, i.e., the current input and previous output, respectively where $j \in [1, N_h]$. Weight matrices $w$ and $u$ and bias vector $b$ are determined during the training phase. The state vector $c_t$ is formed by accumulating $m_t$ over time, as shown in (3). The output vector $h_t$ is formed by applying *tanh* activation function on $c_t$. As shown, the output is related to all previous inputs.

*Long Short-Term Memory (LSTM):* In the above simple RNN cell the effect of all previous information is accumulated in the internal state vector. Gradient-based algorithms may fail when temporal dependencies get too long because gradient values may increase or decrease exponentially [17]. LSTM solves this issue by allowing to forget unessential previous data according to the actual dependencies in the problem. The dependencies are automatically extracted based on the data. This is achieved through forget, input, and output gates [17]. The LSTM cell is shown in Figure 1(b). The gate signals are formed based on $x_t$ and $h_{t−1}$ as shown below.

$$f_t[j] = \sigma\left(\sum_{k=1}^{N_x} w_f[j,k]x_t[k] + \sum_{k=1}^{N_h} u_f[j,k]h_{t-1}[k] + b_f[j]\right) \quad (4)$$

$$i_t[j] = \sigma\left(\sum_{k=1}^{N_x} w_i[j,k]x_t[k] + \sum_{k=1}^{N_h} u_i[j,k]h_{t-1}[k] + b_i[j]\right) \quad (5)$$

$$o_t[j] = \sigma\left(\sum_{k=1}^{N_x} w_o[j,k]x_t[k] + \sum_{k=1}^{N_h} u_o[j,k]h_{t-1}[k] + b_o[j]\right) \quad (6)$$

In the above equations, $\sigma$ denotes the *sigmoid* activation function, and $j \in [1, N_h]$. In the LSTM cell, $m_t$ is computed as before, i.e., as in (1), but (2) and (3) are modified based on the forget, input, and output gate signals as follows:

$$c_t[j] = f_t[j] \times c_{t-1}[j] + i_t[j] \times m_t[j] \quad (7)$$

$$h_t[j] = o_t[j] \times \tanh(c_t[j]) \quad (8)$$

As shown in (7), the forget gate $f_t$ controls carrying of state vector $c$ from time $t − 1$ to time $t$. The input gate $i_t$ adjusts the accumulation of $m_t$ in $c_t$. As shown in (8), the output $h_t$ is formed by applying *tanh* activation function on $c_t$ and is then adjusted by the output gate $o_t$. As the above equations show, the LSTM output would depend on all previous inputs. Previous information is neither completely discarded nor completely carried over to the current state. Instead, the influence of the previous information on the current state is carefully controlled through the gate signals.

## IV. PROPOSED METHOD

In this section, first, we discuss the motivation behind our work and then we present our multi-level binarized neural networks.

### A. Motivation

As mentioned in Section III, LSTM has four distinguished weights including cell memory weight ($w$), forget weight ($w_f$), input weight ($w_i$), output weight ($w_o$). $x_t$, $h_{t-1}$ and $c_{t-1}$ are fed into LSTM cell as inputs for computing cell vector ($c_t$) and the hidden vector ($h_t$) according to Equation (4) to (8). It should be mentioned that all inputs and parameters are full precision in conventional LSTM. Therefore, a huge amount of memory and computation is required.

Operation (4) to (8) are consequently repeated depending on the number of hidden layers in LSTM. In case of complicated applications and to achieve desirable performance, LSTM networks are to become deeper/wider and hence more complex. Therefore, deploying these networks on resource-limited and low-power embedded devices is a major challenge, particularly for extremely power-constrained wearable devices. To cope with this challenge, binarized LSTMs are recently deployed in natural language processing and human activity recognition [19] [20]. For instance, our observations show that by exploiting a binarized LSTM for the IMDB dataset, we have achieved accuracies close to the full-precision ones.

Binarized LSTMs replace MAC operations with bit-wise operations and hence significantly decrease the computations. Also, binarizing both weights and inputs reduces memory storage requirements in binarized LSTMs. However, binarized LSTM suffers from accuracy loss in more complicated applications such as EEG classification. Our observations on EEG classification has demonstrated that binarized LSTM leads to considerable accuracy loss.

On the other hand, wearable devices have limited computation and power resources. Therefore, continuous monitoring on wearable devices requires an accurate as well as light-weight EEG classification algorithm. In fact, our proposed quantization approach is an extension to the state of the art LSTM binarization techniques and not only reduce the computation-intensive MAC operations, but also is suitable for applications in which the conventional binarized LSTMs lead to unacceptable accuracy.

### B. Conventional binarization technique

In binary neural networks, full precision input (I) and weights (w) values are substituted with -1 or +1. Two common methods mostly used for binarizing inputs and weights are called deterministic and stochastic binarization functions according to Equation 9 and 10, respectively [13].

$$x_b = \begin{cases} +1 & if\ x \geq 0, \\ -1 & otherwise \end{cases} \quad (9)$$

$$x_b = \begin{cases} +1 & with\ probability\ p = \sigma(x), \\ -1 & with\ probability\ 1 - p \end{cases} \quad (10)$$

Where x is weight or input and $\sigma$ as a hard-sigmoid function determines the probability, distribution shown in Equation 11. Clip function keeps the $\sigma(x)$ between (0, 1).

$$\sigma(x) = clip\left(\frac{x+1}{2}, 0, 1\right) = max\left(0, \min(1, \frac{x+1}{2})\right) \quad (11)$$

Sign function is a deterministic function and is commonly used for input and weight binarization. This function has a simple hardware implementation compared to stochastic functions. Sign values are encoded in binary such that -1 and 1 are encoded with 0 and 1, respectively, according to (12).

$$sign(x) = \begin{cases} +1\ (mapped\ to\ 1) & if\ x \geq 0, \\ -1(mapped\ to\ 0) & otherwise \end{cases} \quad (12)$$

By using binary values instead of full precision values, dot product operations can just be done by *xnor*-bitcount operations [14]. Figure 2 has shown the relation between the dot product and *xnor*-bitcount operations. It should be noted that employing the scaling factor for the inputs and weights plays a crucial role in decreasing accuracy loss arisen from binarization. Therefore, binarized neural networks mostly use scaling factors for inputs and weights. In [19], the authors have deployed the norm function for achieving the scaling factor. The required computations of this function are complicated and finally, multiplication operations are needed for achieving the result.

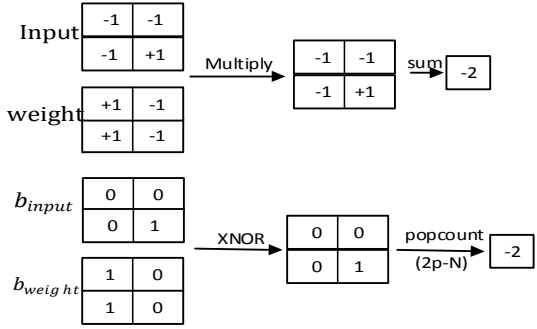

Figure 2. The relationship between the dot product (top) and xnor-bitcount (bottom)

### C. Proposed quantized neural network

Since binarized LSTMs suffer from significant accuracy loss in applications like EEG classification, we proposed a specific multi-level binarization to efficiently quantize both inputs and weights in a way that decreases the number of computations, resources, and memory footprint and yet reduces accuracy loss. To reduce the accuracy loss of the conventional binarized LSTMs we first extend the binarized values to multi-bit fixed-point values and assess the effect of bit width on the accuracy. The results are summarized in Table I.

TABLE I. THE OBTAINED ACCURACY BY CONVENTIONAL FIXED-POINT. "A" (ROWS) AND "W" (COLUMNS) STAND FOR ACTIVATION AND WEIGHT BIT WIDTHS RESPECTIVELY. FP STANDS FOR 32-BIT FULL-PRECISION.

| A/W | 1 | 2 | 3 | 4 | 5 | FP |
|---|---|---|---|---|---|---|
| 1 | 53.92 | 54.24 | 55.06 | 53.59 | 53.16 | 53.92 |
| 2 | 56.86 | 58.49 | 57.51 | 57.51 | 64.37 | 68.62 |
| 3 | 57.51 | 58.49 | 60.78 | 62.74 | 67.48 | 74.34 |
| 4 | 61.43 | 62.17 | 62.58 | 64.05 | 69.6 | 77.77 |
| 5 | 57.35 | 63.56 | 62.74 | 66.99 | 72.22 | 80.88 |
| FP | 62.9 | 67.15 | 66.66 | 78.1 | 81.69 | 82.03 |

As shown in Table I the results of (A, W) = (5,5), sanding for 5-bit for both activations and weights is much lower than the full precision LSTM. To improve accuracy, we would also use the scaling factor which could play a fundamental role in achieving high accuracy. To the best of our knowledge, in the state of the art binarized neural networks, the scaling factors are 32-bit full precision values and therefore impose a 32-bit multiplier to the network. We propose scaling factors that are in the form of powers of two and hence avoid computation-intensive multiplies and exploit simple shift operations instead.

To find the proper values for scaling factors, we have explored the design space to find the appropriate powers of 2 scaling factor. In fact, by constraining the scaling factor to the powers of 2 numbers (like 1, 1/2, 1/4, 1/8, etc.), multiplication can be substituted by the simple shift operation. Based on the obtained scaling factor (α), our proposed multi-level binarization is done according to Algorithm 1.

**Algorithm 1** computing N-level residual binarization

**Inputs**: α
**Output**: $l_1, l_2, \ldots, l_n$

1: $r = x$
2: **for** i = 1 **to** N **do**
3:   $l_i \leftarrow Binarize((Sign(r))$
4:   $r \leftarrow r - Sign(r) \times \alpha/2^{i-1}$
5: **end for**

Where $n, x, l_i$, and $r$ denote the number level of binarization, real value of input or weight, $i^{th}$ level of binarization, and an obtained error arisen from binarization, respectively. The deterministic function Sign(x) is used in each level of the proposed method. By using the scaling factor, x is approximated by $\alpha.sign(x)$ and hence, as show in line 4 of Algorithm 1, the error of binarization is the difference between x and approximated value $\alpha.sign(x)$ which is deployed to determine the next level of binarization. This process is continued based on the desirable number of levels. Note that sign values must be encoded to 0 (-1) and 1 (1), respectively. It should be noted that the primary scaling factor must be halved in the second level and this process is repeated for all levels. To put it simply, in each level, the scaling factor is half of the scaling factor of the previous level. It is worth bearing in mind that the primary scaling factor has obtained through exhaustive explorations on weight or input values.

Using the proposed technique for calculating scaling factors of each level and factorizing the α value, our multi-level binarized values would be as simple as fixed-point values multiplied to α, and hence we can deploy low bit-width fixed-point arithmetic operations which are significantly simplified compared to full precision ones. It should be mentioned that this simple idea has significantly achieved better accuracies than conventional fixed-point which are shown in the experimental results in Section V. As mentioned, the multiplication of the scaling factor is just implemented by a simple shift. We have considered a distinguished scaling factor for each weight, bias, and input based on their value. Therefore, matrix multiplication between weight and input is approximated according to Equation 13.

$$x * w = \alpha_x . ML(x) * \alpha_w . ML(w) \quad (13)$$
$$= \alpha_x * \alpha_w . (ML(x) * ML(w)) = \gamma . (ML(x) * ML(w))$$

Where ML() is our proposed method for multi-level binarization of inputs and weights. $\alpha_x$ and $\alpha_w$ denote the scaling factors of inputs and weights, respectively. Since the scaling factors have primarily obtained by exploration and are constant, we can directly store them and use γ instead of $\alpha_x . \alpha_w$. The number of levels in input and weight can be different in the proposed method. Figure 3 has demonstrated our proposed MAC unit which both weights and inputs must be encoded by our proposed method and the fixed-point MUL unit is deployed to calculate the results. Finally, the obtained results are shifted base on the γ value to achieve the final result. In Figure 3, $X_{ml}$ and $W_{ml}$ denote multi-level binarized input and multi-level binarized weights. LSTM cell comprises four separated weights ($W_c$, $W_f$, $W_i$ and $W_o$) and four separated biases ($b_c$, $b_f$, $b_i$, and $b_o$) as well. All these parameters and $X$ as input are multi-level binarized by our method. It should be mentioned that each weight matrix includes two separate matrixes (recurrent weights and forward weights) which are concatenated together. The first one is multiplied to h and the other one is multiplied to x.

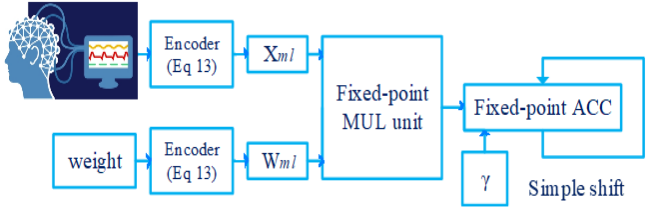

Figure 3. Proposed MAC unit.

## V. EXPERIMENTAL RESULTS

In this paper, we have focused on the EEG dataset, which is not well deployed by binarized LSTM and binary values lead to significant accuracy loss. The number of features and time-steps of the deployed EEG dataset by 3625 samples are 32 and 1300, respectively. Our experiments are presented in two parts. First, we have intelligently explored the design space to find appropriate scaling factors for each parameter. The objective of our proposed method is to achieve the best accuracy. Afterward, we assess the achieved accuracy for various binarization levels, either for weights or inputs. Note that if the number of levels is modified, the primary scaling factor should be changed as well. For more clarification, the primary scaling factor of five various binarization level is demonstrated in Table II. $X$, $W_f$, $W_r$, and $B$ denote input, forward weight, recurrent weight, and bias term, respectively.

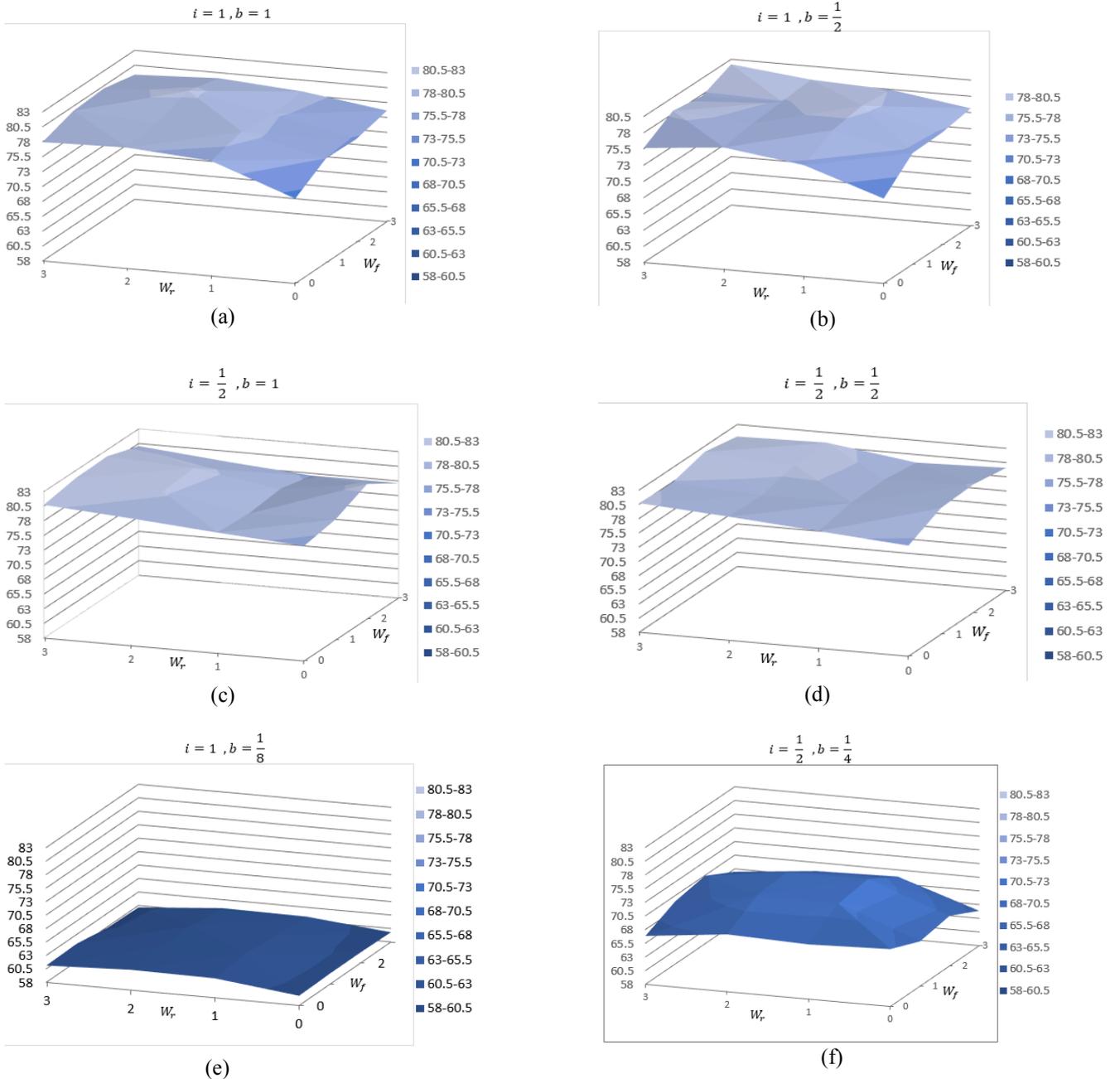

Figure 4. The achieved accuracy with various $W_f$ and $W_r$ by considering input and bias scaling factors as a) (1,1) b) (1,1/2), c) (1/2,1), d) (1/2,1/2) e) (1,1/8) f) (1/2,1/4).

TABLE II. FIVE SAMPLE OF PRIMARY SCALING FACTOR FOR DIFFERENT PARAMETERS.

| Level of binarization (Input, weight) | X | $W_f$ | $W_r$ | B |
|---|---|---|---|---|
| (1,1) | 1/2 | 1/8 | 1/8 | 1/16 |
| (2,2) | 1 | 1/4 | 1/2 | 1 |
| (3,5) | 1/2 | 1/2 | 1/8 | 1/2 |
| (4,5) | 1/2 | 1/2 | 1/8 | 1 |
| (5,5) | 1/2 | 1/4 | 1/8 | 1/2 |

Since the design space of scaling factors for each parameter is a tremendous space, we have intelligently constrained the design space exploration of each parameter's scaling factor based on the parameter's value domain ($X$, $W_f$, $W_r$, $B$). As shown in Table III, we have approximately achieved the same accuracy as full precision accuracy for 5-level binarized weights and inputs. Therefore, for 5-level binarization, we exemplify the method by which we obtain scaling factors of $W_f$ and $W_r$ parameters. We have assumed four specific combinations of suitable input and bias scaling factor. Figure 4 has demonstrated the obtained accuracy with various $W_f$ and $W_r$ scaling factors. Input Scaling factor and bias scaling factor are considered as (1, 1), (1, 1/2), (1/2, 1), and (1/2, 1/2) in Figure 4-a, 4-b, 4-c, and 4-d, respectively.

For more clarification, we have also demonstrated two inappropriate combinations of input and bias scaling factor. As shown in Figure 4, by choosing unsuitable scaling factors, accuracy has dropped 20% to 30% compared to full precision accuracy. As shown in Figure 4, the input scaling factor and bias scaling factor are assumed (1, 1/8) and (1/2, 1/4) in Figure 4-e, 4-f, respectively.

As mentioned in Section IV, conventional fixed-point leads to remarkable accuracy loss, whereas our proposed method achieves close accuracy to full precision ones. Table I and Table III have illustrated the best-obtained accuracies by conventional fixed-point and our proposed method. The comparison between Table I and Table III has shown that our proposed method has significantly improved accuracy (more than 10%) compared to conventional fixed-point. In Table III, 1-level binarization can be considered as the conventional binarized LSTM which leads to significant accuracy loss (~ 22%). As shown in Table III, in comparison with full precision LSTM, we have achieved acceptable accuracies by deploying 3-level and 4-level binarization of inputs and weights, respectively. Moreover, the experimental results have shown that by deploying just 5-level binarization for both inputs and weights, we can obtain the extremely close full-precision accuracy (lower than 0.01% accuracy loss). Therefore, our proposed method has remarkably decreased computations and resources without accuracy loss. Actually, in our proposed method, based on the application constraints the number of bit levels of weights and inputs can be chosen. We have estimated area and delay of the proposed MAC operations versus full precision MAC operations. We have considered the number of gates as an area and logic depth in terms of gate delay as the delay. Figure 5 and Figure 6 have demonstrated the normalized value of the area and the normalized value of the delay of MAC operation based on the maximum value (5-level binarized weights and inputs) for various combinations of ($m$, $n$) level binarized inputs and weights, where m is the activation and n is the weight bit width. Since the area and delay of full-precision (FP) MACs are enormous, we have omitted them and shown in Table IV with accurate multi-level binarized weights and inputs. It should be mentioned that in Table IV, area and delay are normalized based on the full precision area and full precision delay.

TABLE III. THE OBTAINED ACCURACY BY CONVENTIONAL FIXED-POINT. "I" (ROWS) AND "W" (COLUMNS) STAND FOR INPUT AND WEIGHT BIT WIDTHS RESPECTIVELY. FP STANDS FOR 32-BIT FULL-PRECISION.

| I/W | 1 | 2 | 3 | 4 | 5 | FP |
|---|---|---|---|---|---|---|
| 1 | 60.45 | 60.13 | 60.45 | 60.94 | 61.11 | 59.8 |
| 2 | 62.25 | 62.74 | 68.62 | 74.67 | 73.85 | 72.22 |
| 3 | 64.05 | 66.66 | 72.05 | 80.22 | 80.22 | 76.63 |
| 4 | 63.88 | 66.99 | 73.52 | 81.69 | 82.01 | 79.74 |
| 5 | 66.66 | 68.13 | 74.01 | 82.01 | 82.02 | 80.88 |
| FP | 67.48 | 70.75 | 78.28 | 82.019 | 82.0285 | 82.03 |

TABLE IV. THE OBTAINED ACCURACY, THE NORMALIZED VALUE OF MAC AREA AND THE NORMALIZED VALUE OF MAC DELAY FOR ACCURATE MULTI-LEVEL BINARIZED INPUTS AND WEIGHTS AND FULL PRECISION ONE.

| (I/W) | Accuracy | Normalized Delay | Normalized Area |
|---|---|---|---|
| (3,4) | 80.22 | 0.026 | 0.021 |
| (3,5) | 80.22 | 0.035 | 0.029 |
| (4,4) | 81,69 | 0.027 | 0.022 |
| (4,5) | 82.01 | 0.035 | 0.03 |
| (5,4) | 82.01 | 0.035 | 0.03 |
| (5,5) | 82.02 | 0.036 | 0.032 |
| (FP,FP) | 82.03 | 1 | 1 |

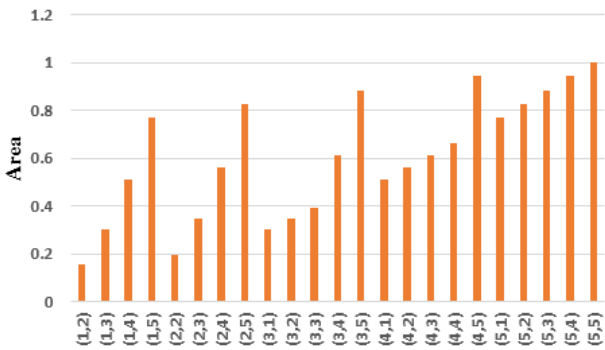

Figure 5. The normalized value of the MAC area based on the maximum value (5-level binarized weights and inputs) for various combinations of (input, weight) bit-widths.

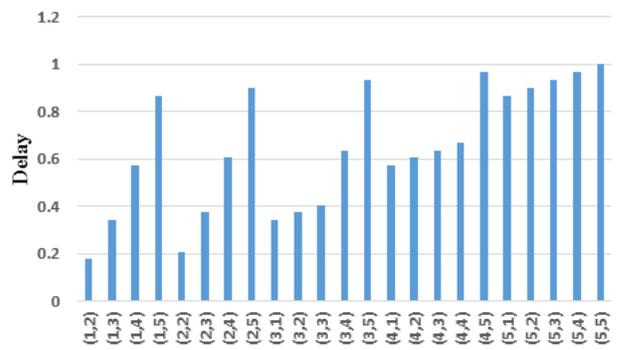

Figure 6. The normalized value of the MAC delay based on the maximum value (5-level binarized weights and inputs) for various combinations of (input, weight) bit-widths.

As shown in Table IV, we have significantly decreased area (~31×) and also delay (~27×) of MAC operations in 65nm technology compared to full precision ones while the same accuracy of full precision accuracy has been achieved on EEG classification by our proposed method (5-level binarization for both inputs and weights).

## VI. CONCLUSION

Binary LSTM is an appropriate approach to realize the deployment of LSTM on low-power embedded devices. However, this approach leads to remarkable accuracy loss for EEG classification which is important to employ in wearable devices. In this paper, a novel multi-level binarized LSTM was proposed for the EEG classification algorithm. Our proposed method not only has significantly decreased computations in terms of area and delay but also has approximately achieved the full precision accuracy for 5-level binarization of inputs and weights. Besides, our proposed method is more accurate compared to fixed-point representation. To achieve more accuracy and still stand low complexity, we explored the power of two distinguished scaling factor for each LSTM parameter and also present simple architecture which is more efficient than fixed point MAC. Moreover, the number of bit-level can be chosen according to applications and their demands in our proposed method.

## ACKNOWLEDGEMENT

KKS has supported this work within the projects DeepMaker and DPAC.